# Unary Coding for Neural Network Learning

Subhash Kak

**Abstract**
This paper presents some properties of unary coding of significance for biological learning and instantaneously trained neural networks.

**Introduction**
Unary code of *n* is generally represented by a string of *n* 1 bits followed by a terminating 0 bit (or equivalently as n 0 bits followed by a terminating 1 bit). Alternative representations for n have *n-1* 1 bits followed by a 0 bit or, if the terminating bit is inessential, by *n* 1 bits. Unary coding has found applications in data compression.

In models of learning inspired by neuroscience, features are often coded spatially so that location denotes number. This is shown in the diagram below.

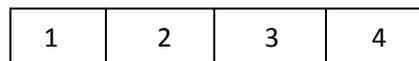

To denote a specific value, one only needs to mark the slot that corresponds to it. This may be done either by marking the specific slot and leaving the others either blank or marked with a different symbol. For example, four different angles are represented in a perceptron learning system by the encodings [1]

   1: 1000
   2: 0100
   3: 0010
   4: 0001

that are termed unary by the authors. This nomenclature is right since the coded information resides in the location of the separation between the 1s and 0s, which in this coding is represented by a single 1.

If this coding is further modified by the rule that all places to the left of 1 should be replaced by 1, we get the familiar:

   1: 1000
   2: 1100
   3: 1110
   4: 1111

Unary representation of small numbers as vertical strokes is attested in the earliest archaeological records in different cultures around the world [2],[3]. The Inca system of counting by knots, called *quipu*, is also essentially a unary system in which the strokes are replaced by the number of knots. Quipus were based on a decimal position system. This unary system, and others elsewhere in the world, evolved with the use of special symbols to represent 5 and 10 and other large numbers.

If unary code words are separately transmitted, then the length of the codeword itself communicates its corresponding *n*, and the distribution of the 0s and 1s in the group is unimportant. In such a case, one can replace the bits by a random distribution of 0s and 1s.

Unary codes are universal codes for symbols whose distribution is monotonic [4]. They have applications in data compression.

In this note we review why unary coding is effective in neural network learning.

The simplest distance measure for binary vectors is the Hamming distance. For such vectors, we would use unary coding if a uniform Hamming distance is required amongst the code words.

In representing points on the plane, binary coding is not uniform in the sense that points far apart can have small Hamming distance and points that are close can have large Hamming distance. Gray coding ensures Hamming distance of one for adjacent elements but the distance is not well defined when we consider elements that are not adjacent.

```
       Number:     1,    2,    3,    4,    5,    6,    7,
Binary coding:  0001, 0010, 0011, 0100, 0101, 0110, 0111 …
  Gray coding:  0001, 0011, 0010, 0110, 0111, 0101, 0100 …
```

For binary coding, the distance between 3 and 4 is 3 bits whereas the distance between 1 and 5 is only 1 bit. Conversely in Gray coding, the distance between 3 and 4 is 1 bit and the distance between 1 and 6 is also 1 bit. Both these mappings are thus unsatisfactory from a distance defining perspective.

Table 1.

| Number | Basic unary | Binary | Gray |
|--------|-------------|--------|------|
| 1 | 1 | 1 | 1 |
| 2 | 11 | 10 | 11 |
| 3 | 111 | 11 | 10 |
| 4 | 1111 | 100 | 110 |
| 5 | 11111 | 101 | 111 |
| 6 | 111111 | 110 | 101 |
| 7 | 1111111 | 111 | 100 |

Although unary coding provides uniform distances, this implicitly assumes that each unary code has been appended with a suitable number of 0s so as to provide the same length for each code word.

**Weight of Unary Codes**

The general unary coding problem is to find a code so that if for numerals $x$ and $y$ the distance between them is given by $d(x-y)$, then

$$d(x-y_1) > d(x-y_2) \text{ if } |x-y_1| > |x-y_2|$$

If the measure used is the Hamming distance, then we have the following corollary regarding weights ($w$) of the unary code:

$$w(y_2) > w(y_1) \text{ if } |y_2| > |y_1|$$

**Application in Neural Networks**

Fixed length unary coding was used in instantaneously trained neural networks to ensure that learning a specific point makes it possible to learn all adjacent (in the Hamming distance sense) points [5]-[9]. It was the uniform property of distance that made instantaneous learning possible since representing the learnt point was straightforward. Once a data point has been recognized, one can achieve generalization by wrapping a region of Hamming radius $r$ units around it.

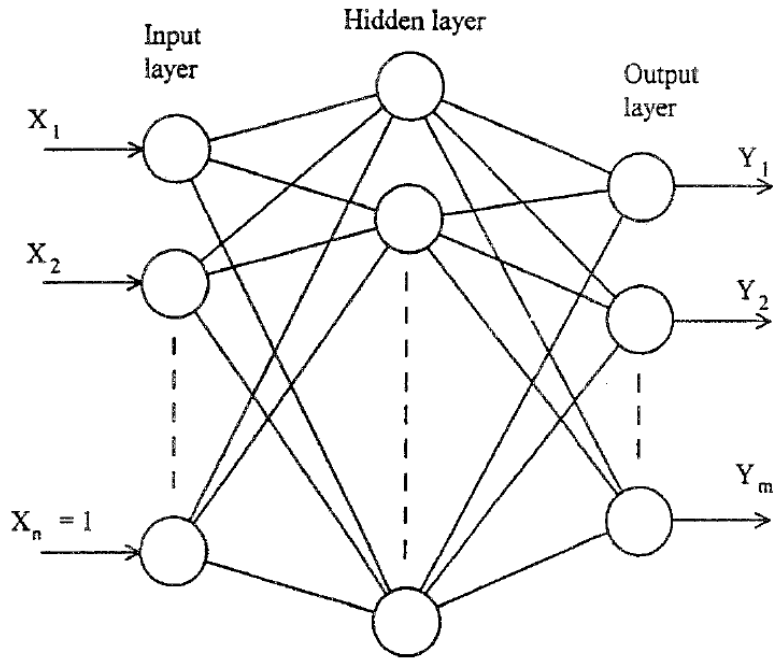

Figure 1. Instantaneously trained neural network

The corner classification (CC) network, which is capable of instantaneous learning, was proposed in various variations [10],[11]. These and its more advanced variants have been implemented in hardware also [12],[13]. There are four versions of the CC technique, represented by CC1 through CC4. The concept of radius of generalization was introduced in CC3 which allowed this neural network to overcome the generalization problem that plagued the earlier CC2 network. Each node in the network acted as a a hyperplane to separate the corner of the *n*-dimensional cube represented by the training vector (hence the name corner-classification, CC, technique).

The CC4 uses a feedforward network architecture consisting of three layers of neurons as shown in Figure 1. The number of input neurons is equal to the length of input patterns or vectors plus one, the additional neuron being the bias neuron, which has a constant input of 1. The number of hidden neurons is equal to the number of training samples, and each hidden neuron corresponds to one training example. The last node of the input layer is set to one to act as a bias to the hidden layer. The binary step function is used as the activation function for both the hidden and output neurons. The output of the activation function is 1 if summation is positive and zero otherwise.

Input and output weights are determined as follows. For each training vector presented to the network, if an input neuron receives a 1, its weight to the hidden neuron corresponding to this training vector is set to 1; otherwise, it is set to -1. The bias neuron is treated differently. If $s$ is the number of 1's in the training vector, excluding the bias input, and the desired radius of generalization is $r$, then the weight between the bias neuron and the hidden neuron corresponding to this training vector is $r - s + 1$. Thus, for any training vector $x_i$ of length $n$ including the bias, the input layer weights are assigned according to the

following equation:

$$w_i[j] = \begin{cases} 1 & \text{if } x_i[j] = 1 \\ -1 & \text{if } x_i[j] = 0 \\ r - s + 1 & \text{if } j = n. \end{cases}$$

The weights in the output layer are equal to 1 if the output value is 1 and –1 if the output value is 0. This amounts to learning both the input class and its complement. The radius of generalization, $r$, can be seen by considering the all-zero input vectors for which $w_{n+1} = r + 1$. The choice of $r$ will depend on the nature of generalization sought. Since the weights are 1, -1, or 0, it is clear that actual computations are minimal. In the general case, the only weight that can be greater in magnitude than 1 is the one associated with the bias neuron.

Several scholars [15],[16],[17] have argued that unary coding is at the basis of fast birdsong learning. This means that there is a neurophysiological justification for the unary mapping of instantaneously trained neural networks.

For the uniform Hamming distance requirement, unary coding may be performed as below:

Table 2.

| n | Unary code | Fixed length unary code |
|---|---|---|
| 0 | 0 | 0000000000 |
| 1 | 10 | 0000000001 |
| 2 | 110 | 0000000011 |
| 3 | 1110 | 0000000111 |
| 4 | 11110 | 0000001111 |
| 5 | 111110 | 0000011111 |
| 6 | 1111110 | 0000111111 |
| 7 | 11111110 | 0001111111 |
| 8 | 111111110 | 0011111111 |
| 9 | 1111111110 | 0111111111 |
| 10 | 11111111110 | 1111111111 |

The fixed length unary code suffers from a disadvantage when it is used in communications problems because it does not have any error correction capability in it. In order to endow the codes with the ability to combat errors, it is essential to increase the minimum Hamming distance between code words beyond the 1 that we have in the mapping of Table 2.

A straightforward way of increasing Hamming distance between code words is to code *n* by a string of *kn* 1 bits followed by a terminating 0 bit (or equivalently as n 0 bits followed by a terminating 1 bit). This will ensure that the minimum Hamming distance between code words is *k-1*. This is not particularly efficient. This, or some other generalized unary coding scheme, may actually be in use in biological systems so as to incorporate error-correction.

**Conclusions**

Generalizations of unary coding may be obtained by considering the minimum code word length to besome integer, and by considering code words that have been made roughly equal in the number of 0s and 1s.

If unary code words are separately transmitted, then the length of the codeword itself communicates its corresponding *n*, and the distribution of the 0s and 1s in the group is unimportant. In such a case, one can replace the bits by a random distribution of 0s and 1s and choose from a variety of such sequences [18],[19],[20].